# General solutions for nonlinear differential equations: a rule-based self-learning approach using deep reinforcement learning

Shiyin Wei[a,b,c,] †, Xiaowei Jin[a,b,c,] †, Hui Li[a,b,c,]∗

[a] *Key Lab of Smart Prevention and Mitigation of Civil Engineering Disasters of the Ministry of Industry and Information Technology, Harbin Institute of Technology, Harbin 150090, China.*

[b] *Key Lab of Structures Dynamic Behavior and Control of the Ministry of Education, Harbin Institute of Technology, Harbin 150090, China.*

[c] *School of Civil Engineering, Harbin Institute of Technology, Harbin 150090, China.*

†These authors contributed to the work equally and should be regarded as co-first authors.

*Corresponding author. E-mail address:* lihui@hit.edu.cn (H. Li).

**ABSTRACT**

A universal rule-based self-learning approach using deep reinforcement learning (DRL) is proposed for the first time to solve nonlinear ordinary differential equations and partial differential equations. The solver consists of a deep neural network-structured actor that outputs candidate solutions, and a critic derived only from physical rules (governing equations and boundary and initial conditions). Solutions in discretized time are treated as multiple tasks sharing the same governing equation, and the current step parameters provide an ideal initialization for the next owing to the temporal continuity of the solutions, which shows a transfer learning characteristic and indicates that the DRL solver has captured the intrinsic nature of the equation. The approach is verified through solving the Schrödinger, Navier–Stokes, Burgers', Van der Pol, and Lorenz equations and an equation of motion. The results indicate that the approach gives solutions with high accuracy, and the solution process promises to get faster.

**Keywords**: nonlinear differential equations; rule-based solving method; deep reinforcement learning; general solution; transfer learning

| Nomenclature | | $\mathcal{S}$ | Domain of the state |
|---|---|---|---|
| DRL | Deep reinforcement learning | $a$ | Action |
| ODE | Ordinary differential equation | $\mathcal{A}$ | Domain of the action |
| PDE | Partial differential equation | $\theta$ | Trainable parameters of the policy network |
| AGI | Artificial general intelligence | $\pi_\theta(a\|s)$ | Probabilistic policy of the action given the state |
| DNN | Deep neural network | $\mu_\theta$ | Mean value of the action determined by the probabilistic policy of the action given the state |
| MDP | Markov decision process | $\sigma_\theta$ | Standard deviation of the action determined by the probabilistic policy of the action given the state |
| SDOF | Single degree of freedom | $r(s,a)$ | Imbalance of an ODE or PDE |
| MDOF | Multi-degree of freedom | $r_{Eq}$ | Imbalance of governing equation |
| $\mathbf{x}$ | Spatial coordinate | $r_B$ | Imbalance of boundary conditions |
| $t$ | Time coordinate | $r_I$ | Imbalance of initial conditions |
| $u$ | Solution of the ODE or PDE | $J$ | Loss function of policy network |
| $u_t$ | Derivative of $u$ with respect to time | $\mathrm{E}_{\pi_\theta}$ | Expectation calculated on probabilistic policy |
| $\vartheta$ | Parameters of ODE or PDE | $\hat{u}$ | Candidate solution sampled from policy |
| $s$ | State | $N$ | Batch size of sampling states |

# 1. Introduction

Differential equations, including ordinary differential equations (ODEs) and partial differential equations (PDEs), form a description of the dynamical nature of the world around us. General solutions of differential equations form the basis on which artificial general intelligence (AGI) agents are able to understand the physical world. However, both analytical and numerical methods require strict technical training [1-3], and this limits the capability of a more intelligent computer. Therefore, there is a need for a simpler and more general solution that allows currently available computers to learn from the training process of a network (or "learning to solve equations") [4, 5].

Deep learning and reinforcement learning [6, 7], which have undergone rapid development in recent decades, are of some use in realizing this idea. Data assimilation methods using deep learning [8, 9] have recently been proposed to encode the Navier–Stokes equations in a neural network to predict a variety of quantities of interest. Although solutions of the Navier–Stokes equations can thereby be obtained, these methods still require a large number of measurements of the flow field. There are also some other deep-learning-based methods, which can be categorized into three types: (1) those that map directly to the solution represented by a deep neural network (DNN) in a continuous manner that is similar to the analytical solution [10, 11] and in which the data used to train the network are randomly sampled within the entire solution domain in each training batch, including initial and boundary conditions; (2) those that map directly to the solution in a discretized manner that is similar to the numerical solution [12-14]; and (3) those that map indirectly to the internal results or parameters of a numerical solution and use these internal results to derive the numerical solution [11, 15]. The essential feature of all of these methods is that they take advantage of the nonlinear representation capabilities of DNNs. Recent progress [16] in mechanics using these capabilities has been reported. For example, Li et al. [17] have developed a generative adversarial network model to map latent variables to microstructures and have used it in material design.

The solutions are either directly output by the neural network or numerically derived from the outputs of the network, and the solution task is regarded as a weak-label task, with the governing equation being treated as the weak label to calculate the loss function of the network. The term "weak label" emphasizes the difference from the label in supervised learning; i.e., the true solutions are not known in these tasks, instead, when we get a candidate solution from the neural network output, we can tell how far the output solution is to the true solution by the imbalance of the governing equation. Because of the weak-label property, a solution obtained using deep learning may be unstable for high-dimensional ODEs/PDEs with local optima. Also, when equations are solved in the whole time domain, a long computational time may be necessary or the procedure may even fail. This problem arises because there is no exploitation of possible network parameter transfers among time steps. Hence, we propose here a deep reinforcement learning (DRL) paradigm for solution of ODEs/PDEs. DRL with exploration capability is naturally suited to weak-label tasks by a trial-and-error learning mechanism that can eliminate trapping in local optima [7, 18]. Taking the game of Go for example [19], the only prior information about the task is the playing rules that define win or lose, and the label (or score) of each step is given by whether there is a win or a loss after the whole episode of playing rather than by an exact score. Therefore, a huge number of optima exist in the action space, and a successful agent playing the game using DRL consists of a policy network that provides candidate playing actions with a probability distribution and a value network that provides the expected return of the action. The exploration attribute enables the agent to explore a larger region of action space and makes it more likely that the agent will escape from local optima. Moreover, examples using DRL in mechanism and materials have been reported in [20, 21], where a meta-modeling framework has been proposed for the improvement of constitutive models and boundary problems have been discussed.

When employing reinforcement learning, we are essentially treating the solution of differential equations as a control task. The state (or observation in some of the literature) is the given solution domain, and the action is the solution of an equation in this domain, so the goal is to find a proper action to balance the governing equation within an acceptable error. A DNN structured policy network (inspired by physics-informed neural networks [11] ) is used to output the action policy given a state, and candidate actions are sampled from the policy. The governing equation is embedded in the network as a critic. The gradients of the policy network are then calculated based on the critic calculated when candidate actions are substituted into the equation. It is worth noting that no data about the solutions are required in the DRL approach, which is thus not only very different from data-driven models [22] and data assimilation methods [8, 9] using the deep learning technique, but also more flexible because data are very hard to obtain in some scenarios. Therefore, the proposed approach can be categorized as a rule-based methodology. In this study, we employ a discretized formalization in the time domain, and the solution at each time step is trained in a reinforcement learning paradigm, so the parameters of the policy network at the current time step provide an ideal initialization for the next step, i.e., they have transfer learning characteristic. As a beneficial result of this transfer characteristic, the solution process promises to become be faster.

The major contribution is summarized as: an alternative method for solving differential equations by the DRL approach is proposed, which is a rule-based self-learning method that requires the least human built-in knowledge. The structure of the remainder of this article is as follows. In Section 2, the structure of the DRL framework is established, and the rules for setting the loss functions and training are described in detail. Section 3 presents solutions of various nonlinear differential equations obtained using the DRL framework, all of which are found to agree well with those obtained by high-order numerical methods or analytical solutions, with the exception of the Lorenz equations with a parameter set causing a chaotic response, which shows that the DRL solver performs well for equations with stable solutions but fails for equations with unstable solutions. In Section 4, the transfer learning characteristic that accelerates the solution process is revealed, and the limitations of the DRL solver for equations with unstable solutions (e.g., chaos) are discussed. Finally, a summary is provided in Section 5.

## 2. Methodology

### 2.1 Deep Reinforcement Learning framework

A general nonlinear differential equation can be written in the form

$$u_t + \mathcal{N}(u,\vartheta) = 0 \qquad (1)$$

where $u(\mathbf{x},t)$ denotes the latent solution of the equation, $u_t$ is the derivative with respect to time, and $\mathcal{N}(\cdot,\vartheta)$ is a nonlinear operator parameterized by $\vartheta$. Given the necessary conditions, i.e., the initial and boundary conditions, the solution of this equation is a nonlinear mapping from the solution domain $(\mathbf{x},t)$ to the solution $u(\mathbf{x},t)$. DNNs have shown remarkable success in the learning of high-dimensional nonlinear functions [22, 23]. The proposed DRL method approximates this nonlinear mapping with a probabilistic policy $\pi_\theta(a|s)$ implemented using a DNN parameterized by $\theta$. The candidate solution can be sampled from the probability policy, i.e., $\hat{u} \sim \pi_\theta(a|s)$, where $s$ in terms of state (in DRL) is the sample point from the solution domain and $a$ in terms of action (in DRL) is the candidate solution. Therefore, the policy network takes the sampling points $S:\{s_i = (\mathbf{x}_i, t_i)\}$ $i = 1, \cdots, N$ from the continuous solution domain $\Omega \in \mathcal{S}$ as inputs and then outputs the statistics (the mean value $\mu_\theta(s_i)$ and standard deviation $\sigma_\theta(s_i)$) to construct the probabilistic policy $\pi_\theta(a|s_i)$ in Gaussian form. The candidate solution is therefore sampled from the policy $\hat{u}_i(\mathbf{x}_i, t_i) \sim \pi_\theta(a|s_i)$.

The intuitive basis for employing DRL to solve differential equations is that the procedure of solving equations can be dealt with as a trial-and-error process consisting of two steps: first guess a candidate solution $a \in \mathcal{A}$ from the computational domain under the given state $s \in \mathcal{S}$; then criticize and improve it by calculating the loss function using the governing equation and the initial and boundary conditions, where $\mathcal{A}$ and $\mathcal{S}$ are the continuous domains of the solution. The governing differential equation together with its boundary and initial conditions are embedded in the loss function as a rule for solving differential equations, the imbalance $r(s,a)$ (including the general imbalance of the governing equation $r_{Eq} = -(\hat{u}_t - \mathcal{N}(\hat{u}, \vartheta))^2$ and the imbalances of the boundary and initial conditions $r_B$ and $r_I$) is taken as the critic, and the goal of the solving task is to find a proper policy that makes $r(s,a)$ a minimum or within the error threshold.

The DRL solver contains a DNN structured solution policy and a rule-based critic. The policy network inputs the sampled points set from the computational domain and outputs the corresponding probabilistic policy of the solutions on the points set, candidate solutions are sampled according to the policy (the output of the DNN), and then the critic is used to calculate the loss of the policy network (i.e., the imbalance of the equation and the initial and boundary conditions) by a given rule that reflects the characteristics of the differential equation. In the solution process, we set time in a discretized format in which the DRL solver treats the solution of the differential equation as a set of multiple tasks that share the same governing equation rule.

Each task provides the temporal solution of the nonlinear differential equations at that time step, and the loss threshold of the current step must be satisfied before the solution proceeds. Because the derivative is very important in ODEs and PDEs, the discretized time step is most prone to numerical errors. Specifically, if the discretized time step is large, the accuracy of the derivative will decrease, inducing further errors in the solution, even to the extent that it no longer converges.

From physical considerations, the boundary and initial conditions are continuous in time and therefore so are the parameters of the policy network: the parameters at the current step are used as an ideal initialization for the next step, which enables the parameters of the DNN to transfer with the time step in the solution process, which therefore becomes faster for the remaining time steps. Partial derivatives in equations are calculated based on the gradient of the output $\mu_\theta$ of the network with respect to the input $(\mathbf{x}_i, t_i)$; to preserve the continuity of these derivatives, the activation functions of the hidden layers are chosen as tanh, rather than the ReLU function. The policy network is trained based on the calculated loss using gradient descent (the Adam method in this study); to preserve the gradient of the policy network, the candidate solution is sampled on the mean value $\mu_\theta$ of the network's output.

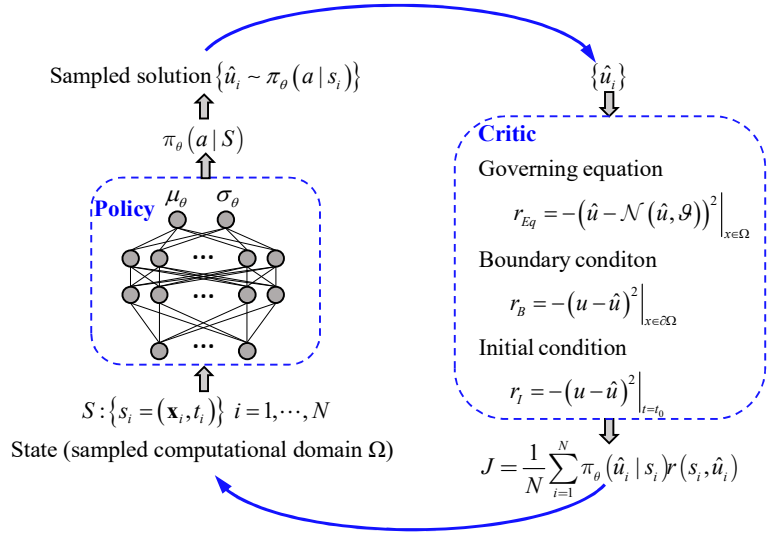

**Fig. 1** DRL framework for equation solution: the action is defined as the candidate solution of the differential equation and is sampled from the output policy of the policy network, and the state is defined as the sampled point from the continuous solution domain of the differential equation and is the input to the policy network. The imbalance of the governing equation under the given boundary and initial conditions when the candidate solutions are substituted provides the critic and the feedback to the policy network.

As balancing of the governing equation is the objective, the solving task can be handled as a one-step Markov decision process (MDP). The loss function of the policy network over the continuous solution action space is written as (for convenience, the symbols used here are those familiar from classic reinforcement learning; the correspondence to a specific solution problem can be found from Fig. 1):

$$J(\theta) = \mathrm{E}_{\pi_\theta}[r] = \int_{\mathcal{S}} ds \int_{\mathcal{A}} \pi_\theta(a|s) r(s,a) da \tag{2}$$

Silver et.al [24] showed an outperformed power of the deterministic version of the policy gradient (DPG) that requires fewer samples to train. In this approach, actions are directly sampled as mean values of the Gaussian distribution $a = \mu_\theta(s_i)$ to preserve gradient backpropagation to $\theta$. Thus, the policy gradient is rewritten in the context of Monte Carlo sampling as

$$\nabla_\theta J(\theta) \approx \frac{1}{N} \sum_{i=1}^{N} \nabla_\theta \pi_\theta\left(\mu_\theta(s_i) | s_i\right) r\left(s_i, \mu_\theta(s_i)\right) \tag{3}$$

where $N$ is the batch size of the sampling states $s_i = (\mathbf{x}_i, t_i)$ $i = 1, \cdots, N$, and the weight $\pi_\theta(s_i, \mu_\theta(s_i))$ provides the likelihood under the candidate action $\mu_\theta(s_i)$ to enhance the loss critic of the candidate action to avoid local optima. The rule-based DRL approach treats the discretized time solution as multi-tasks that share the same governing equation. An individual DRL agent is trained for the solution at each time step rather than using a pre-trained model to generalize another solution case in the data-driven approaches [22]. Therefore, the overfitting problem, which is usually used to depict the poor generality owing to the complexity of the machine learning model and the parameters in regression and classification problems, is not encountered in this study.

**2.2 Setting of DRL hyperparameters**

Adam (adaptive moment estimation) [25] is employed as the optimization algorithm for training the policy network, in which learning rate is an essential hyperparameter. A decaying learning rate is used to train the network; the high learning rate in the early training stage makes a rapid exploration of parameter space and the later low learning rate makes the exploration more stable. Meanwhile, the selection of activation functions, the number of hidden layers, the number of nodes of each hidden layer in the deep policy network and the discretized time-step for the discretized form of DRL approach are also hyperparameters to be set. The way to guarantee success in the present study is continuously increasing the number of hidden layers/nodes until the stable solution is obtained [26]. All the hyperparameters are given in companion with the simulated examples. In the DRL solver, the derivatives in the ODEs/PDEs are approximated by the derivates of the output with respect to the input of the policy network. Hence, the activation function needs to be differentiable. Though suffering from vanishing gradient, a sigmoidal activation function with continuous derivates is used in the solver. Because the hyperbolic tangent activation function tanh is similar to the identity function near 0, so typically it performs better than the logistic sigmoid [26]. Therefore, tanh is chosen as the activation function in this solver. In addition to this hyperparameter setting methodology, all the hyperparameters can also be optimized by grid or random search [27] or Gaussian processes [28]. Both manual hyperparameter tuning and algorithm-based hyperparameter tuning are aimed at obtaining a solution, and both approaches can be used to determine the hyperparameters.

## 3. Results

This section demonstrates the implementation of the DRL framework and the solutions of some well-known ODEs and PDEs that are thereby obtained.

**3.1 Van der Pol equation**

The Van der Pol equation is a second-order differential equation describing an oscillator with nonlinear damping [29]:

$$x_{tt} - \alpha(1-x^2)x_t + x = \beta p(t) \tag{4}$$

where $x_{tt}$ is the second-order derivative, $\alpha, \beta > 0$ are scalar parameters, and $p(t)$ is the external excitation. A high-order differential equation can always be rewritten in a state representation consistent with Eq. (1) in this case as

$$\begin{aligned} x_t &= y \\ y_t &= \alpha(1-x^2)y - x + \beta p(t) \end{aligned} \tag{5}$$

Therefore, in the following, we present equations in their most well-known high-order forms. In solving the Van der Pol equation, the temporal interval $\Delta t$ is set as 0.001 s, the input of the policy network of each discretized step $i$ is a minibatch of $s_i = (t_i)$ where $t_i \in [0, \Delta t]$, and the batch size is set as $N = 100$. The DNN is constructed as the policy network, a three-hidden-layer DNN with 32 nodes in each layer is employed, the output $\mu_\theta(s_i) = [\mu\_x_i, \mu\_y_i]$ consists of two nodes that generate candidate solutions $\hat{x}_i = (t_i + 1) \cdot \hat{x}_{i-1} + t_i \cdot \mu\_x_i$ and $\hat{y}_i = (t_i + 1) \cdot \hat{y}_{i-1} + t_i \cdot \mu\_y_i$ to naturally satisfy the initial conditions [10], the derivatives $\hat{x}_{ti}$ and $\hat{y}_{ti}$ are set as the derivative of the network, and the reward function is deduced from Eq. (5) as:

$$r(t_i) = -(\hat{x}_{ti} - \hat{y}_i)^2 - (\hat{y}_{ti} - \alpha(1 - \hat{x}_i^2)\hat{y}_i + \hat{x}_i - \beta p(t))^2 \quad (6)$$

where the external excitation $p(t) = \cos(\omega t)$ and the error threshold is set as $10^{-4}$.

Figure 2a presents the results together with those from the explicit Runge–Kutta (4, 5) method (the ODE45 method), and it can be seen that the results agree well with each other [30]. This shows that the DRL approach is effective in solving the Van der Pol equation.

### 3.2 Equation of motion

Equation of motion is frequently used to describe structural or system dynamics [31]. The nonlinear form with Bouc-Wen hysteresis model for a single degree of freedom (SDOF) system and a multi-degree of freedom (MDOF) system are

$$\begin{aligned} mx_{tt} + cx_t + \alpha k x + (1-\alpha)kz &= p(t) \\ z_t &= Ax - \beta z |x_t||z|^{n-1} - \gamma x_t |z|^n \end{aligned} \quad (7)$$

$$\begin{aligned} MX_{tt} + CX_t + \alpha K_1 X + (1-\alpha)K_2 Z &= P(t) \\ X = (x^{(1)}, x^{(2)}, \cdots, x^{(N)}) \quad Z &= (z^{(1)}, z^{(2)}, \cdots, z^{(N)}) \\ z_t^{(j)} &= A y_t^{(j)} - \beta z^{(j)} |y_t^{(j)}||z^{(j)}|^{n-1} - \gamma y_t^{(j)} |z^{(j)}|^n \quad (j = 1, 2, \cdots, N) \\ y^{(1)} &= x^{(1)} \\ y^{(j)} &= x^{(j)} - x^{(j-1)} \quad (j = 2, 3, \cdots, N) \end{aligned} \quad (8)$$

where $m, c, k$ ($M, C, K$) are the mass, damping, and stiffness matrix of the structure in the SDOF (MDOF) systems, respectively, $N$ is the degree of the MODF system, $p(t)$ ($P(t)$) is the external excitation, $\alpha kx, (1-\alpha)kz$ ($\alpha K_1 X, (1-\alpha)K_2 Z$, where $K_1, K_2$ are the linear and nonlinear stiffness matrices) are the linear and nonlinear resiliences, respectively, $A, \beta, \gamma, n$ are the Bouc-Wen parameters that control the shape and size behavior of the hysteresis model, $y^{(j)}$ and $z^{(j)}$ are the inter-layer displacement and resilience, and $y_t^{(j)}, z_t^{(j)}$ are the derivative of $y^{(j)}$ and $z^{(j)}$, respectively.

Taking the three-degree-of-freedom system as an example, the structural parameters are

$$M = \begin{pmatrix} m^{(1)} & & \\ & m^{(2)} & \\ & & m^{(3)} \end{pmatrix} \quad C = \begin{pmatrix} c^{(1)} + c^{(2)} & -c^{(2)} & \\ -c^{(2)} & c^{(2)} + c^{(3)} & -c^{(3)} \\ & -c^{(3)} & c^{(3)} \end{pmatrix}$$

$$K_1 = \begin{pmatrix} k^{(1)} + k^{(2)} & -k^{(2)} & \\ -k^{(2)} & k^{(2)} + k^{(3)} & -k^{(3)} \\ & -k^{(3)} & k^{(3)} \end{pmatrix} \quad K_2 = \begin{pmatrix} k^{(1)} & -k^{(2)} & \\ & k^{(2)} & -k^{(3)} \\ & & k^{(3)} \end{pmatrix}$$

$$m^{(1)} = m^{(1)} = m^{(1)} = 1 kg \quad c^{(1)} = c^{(2)} = c^{(3)} = 2 N \cdot s / m \quad k^{(1)} = k^{(2)} = k^{(3)} = 100 N / m$$

$$P(t) = -M \left( \ddot{u}_g, \ \ddot{u}_g, \ \ddot{u}_g \right)^T$$

$$\alpha = 0.1, \ A = 1, \ \beta = 0.5, \ \gamma = 0.05, \ n = 1$$

where $\ddot{u}_g$ is an external earthquake excitation, and is chosen here as the El Centro signal as illustrated in Fig. 2b.

The policy network has the same structure as in the case of the Van der Pol equation: three-hidden-layer DNN with 32 nodes in each layer is employed, and the output $\mu_\theta(s_i) = \left[ \mu\_x_i^{(1)}, \mu\_x_i^{(2)}, \mu\_x_i^{(3)}, \mu\_y_i^{(1)}, \mu\_y_i^{(2)}, \mu\_y_i^{(3)}, \mu\_z_i^{(1)}, \mu\_z_i^{(2)}, \mu\_z_i^{(3)} \right]$ consists of 9 nodes, with a similar trick to that adopted in the Van der Pol equation case being used to naturally satisfy the initial conditions. The temporal interval $\Delta t$ is set as 0.01 s, and the input of the policy network of each discretized step $i$ is a minibatch of $s_i = (t_i)$, where $t_i \in [0, \Delta t]$, and the batch size is set as $N = 100$. The error threshold of the critic is set as $10^{-3}$. Figures. 2b and 2c compare the DRL solution results with those of the ODE45 method.

### 3.3 Lorenz equations

The Lorenz equations are

$$\begin{cases} x_t = \sigma(y-x) \\ y_t = \rho x - y - xz \\ z_t = -\beta z + xy \end{cases} \quad (9)$$

where $\sigma$ (the Prandtl number), $\rho$ (the Rayleigh number), and $\beta$ (an aspect ratio) are the three parameters of the system. In particular, for fixed $(\sigma, \beta) = (10, 8/3)$, the system becomes chaotic when $\rho > \rho_H \approx 24.74$. The solution has the symmetric property such that if $(x(t), y(t), z(t))$ is a solution, then so is $(-x(t), -y(t), z(t))$. Specifically, the solution $u_1(t) = (x_1(t), y_1(t), z_1(t))$ starting from the initial state $[0, 2, 0]$ and the solution $u_2(t) = (x_2(t), y_2(t), z_2(t))$ starting from the initial state $[0, -2, 0]$ should be symmetric with respect to the Z-axis.

Two cases of the Lorenz equation starting from the initialization $(0, 2, 0)$ are investigated here, namely, that with parameter setting $\sigma = 10, \beta = 8/3, \rho = 15$, which has a stable solution, and that with $\sigma = 10, \beta = 8/3, \rho = 28$, which has an unstable solution exhibiting chaotic behavior. The same DNN for the policy network is employed, with three hidden layers with 32 nodes in each layer. The output is set to be $\mu_\theta(s_i) = [\mu\_x_i, \mu\_y_i, \mu\_z_i]$ and the temporal interval $\Delta t$ is set as 0.001 s. The time derivatives of $(x_t, y_t, z_t)$ are handled in a similar way as in the solution of the Van der Pol equation by using the derivatives of the network. The results are illustrated in Figs. 2d and 2e. The solutions for DRL and ODE45 show good consistency for the stable solution but inconsistency for the unstable chaotic solution; the limitations encountered by the DRL in the chaotic case are discussed in Section 4.2.

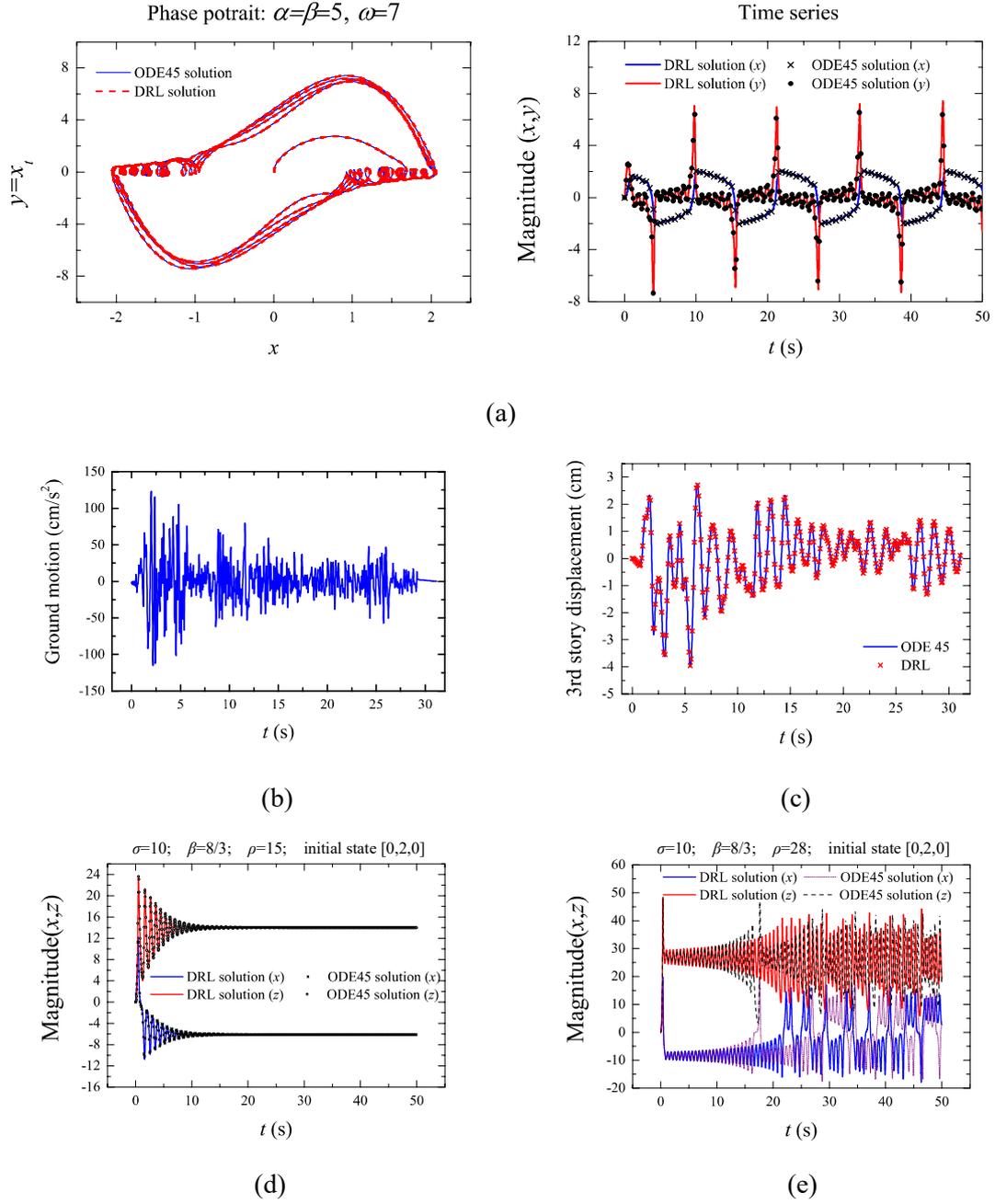

**Fig. 2** Comparison between DRL solutions and those obtained using the Runge–Kutta (4, 5) method (ODE45): (a) solutions of the Van der Pol equation; (b) excitation input signal; (c) solutions of the equation of motion; (d) stable solution of the Lorenz equations; (e) unstable chaotic solution of the Lorenz equations.

### 3.4 Burgers' equation

Burgers' equation arises in a wide range of applications, such as nonlinear acoustics and gas dynamics [32]. The general form of Burgers' equation is,

$$\frac{\partial \mathbf{u}}{\partial t} + (\mathbf{u} \cdot \nabla)\mathbf{u} - \nu\nabla^2\mathbf{u} = 0 \tag{10}$$

where $\mathbf{u}(\mathbf{x},t)$ is a given field and $\nu$ is the diffusion coefficient.

The diffusion coefficient is a critical parameter that influences the solution topology of Burgers' equation, and first of all we consider three values of the diffusion coefficient: 0.1, 0.05, and 0.01. Taking the one-dimensional Burger's equations as an example, the diffusion coefficient, the computational domain, and the initial and boundary conditions are as follows:

$$\nu = 0.1, 0.05 \text{ and } 0.01, \ x \in [-1,1], \ t \in [0,1],$$

$$u(x,0) = -\sin(\pi x), \ u(-1,t) = u(1,t) = 0$$

To let the policy network automatically satisfy the initial and boundary conditions, the policy network in the DRL framework is written as,

$$\hat{\mathbf{u}}(\mathbf{x},t) = t(x+1)(x-1)f(\mathbf{x},t;\theta) - \sin(\pi x)$$

where $\theta$ is the trainable variable in the policy network. Then, the terms in Burgers' equation can be derived as

$$\frac{\partial \mathbf{u}}{\partial t} = \frac{\partial f}{\partial t}, \ \frac{\partial \mathbf{u}}{\partial x} = \frac{\partial f}{\partial x}, \ \frac{\partial^2 \mathbf{u}}{\partial x^2} = \frac{\partial^2 f}{\partial x^2}$$

The critic for each time step is

$$r_{Eq} = -\frac{1}{N_{Eq}}\sum\left(\frac{\partial \hat{\mathbf{u}}}{\partial t} + (\hat{\mathbf{u}} \cdot \nabla)\hat{\mathbf{u}} - \nu\nabla^2\hat{\mathbf{u}}\right)^2 \tag{11}$$

The loss function can then be calculated using Eq. (2).

A DNN consisting of seven hidden layers is used for inferring the policy. The temporal interval $\Delta t$ is set as 0.01, and the input of the policy network at each discretized step $i$ is a minibatch of $s_i = (t_i, x_i)$. The output of the network includes $\mu_\theta(s_i) = \mu_u$ and the standard deviation $\sigma_u$. The number of nodes in each hidden layer is [32 64 64 64 64 64 32], and the activation function is tanh. At each time step, 99 points are sampled for the current step and $99i$ points for the previous step to form the $s_i = (t_i, x_i)$. The learning rate is set to exponential decay, with an initial value of $5 \times 10^{-4}$ and a decay rate of 0.98 (decaying every 100 steps, but no less than $10^{-5}$). After iteration, the mean square error $r_{Eq}$ converges to less than $5 \times 10^{-5}$ for all values of the diffusion coefficients at each time step. A comparison between the DRL solutions and the solutions obtained by Cole [1] is shown in Fig. 3. It should be noted that 50 terms are taken in the infinite-series solution. The DRL solutions, i.e., the spatial–temporal cloud map of the solution and the solutions selected at some spatial and temporal points, agree well with Cole's results. As shown in Fig. 3c, a shock wave appears as $\nu$ decreases, and the absolute value of the first derivative is quite large. However, the DRL solutions are able to accurately capture the shock wave.

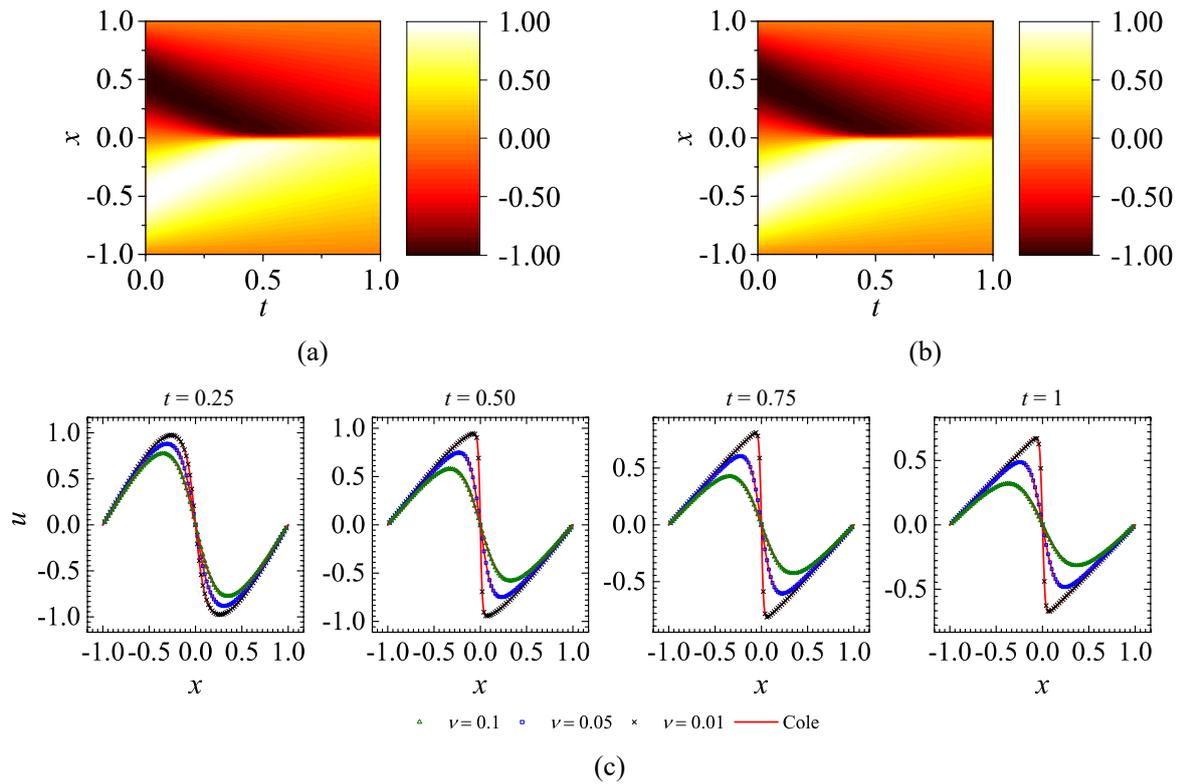

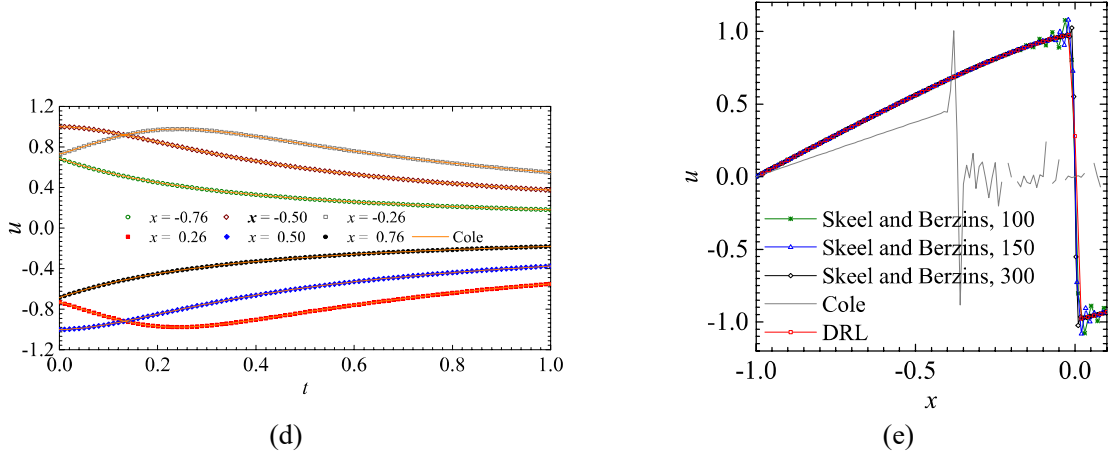

(d)   (e)

**Fig. 3** Solutions of the one-dimensional Burgers' equation: (a) spatial–temporal solution by Cole for $\nu = 0.01$; (b) spatial–temporal DRL solution for $\nu = 0.01$; (c) comparison between the DRL solution and the exact solution at four selected temporal snapshots for various diffusion coefficients; (d) comparison between the DRL solution and the exact solution at six selected spatial points for $\nu = 0.01$; (e) comparison of solutions of the one-dimensional Burgers' equation for $\nu = 0.002$ at $t = 0.55$.

To investigate further the ability of the DRL approach to capture the shock wave, an example with a low diffusion coefficient $\nu = 0.002$ is studied, and the local results at $t = 0.55$ are shown in Fig. 3e. As can be seen, the DRL approach finds a stable shock wave, whereas the analytical solution formulated in series form by Cole [1] fails to give a stable solution with double-precision floating point value. This is a consequence of the subtraction of two large numbers in the denominator of the analytical solution. In addition, a Galerkin finite element method by Skeel and Berzins [33], which is coded in the pdepe function in MATLAB, is implemented to obtain a numerical solution, which is also shown in Fig. 3e. It is found that this numerical method fails to give a stable solution with 100 and 150 uniform grids in space, although it can effectively capture the shock wave for all the examples above. When 300 uniform grids in space are set, this numerical method reaches a stable state, and the DRL solution agrees well with this result. Therefore, the DRL approach has a strong ability to capture the shock wave.

### 3.5 Schrödinger equation

The Schrödinger equation describes the changes in the quantum state of a physical system over time [34]. The nonlinear Schrödinger equation considered in this study is,

$$\frac{\partial u}{\partial t} = 0.5i \frac{\partial^2 u}{\partial x^2} + i|u|^2 u \qquad (12)$$

The computational domain and the initial and boundary conditions for the Schrödinger equation are set as follows:

$$x \in [-5,5], \; t \in [0, \pi/2]$$
$$u(x,0) = 2\operatorname{sech}(x), \; u(t,-5) = u(t,5)$$
$$\left.\frac{\partial u}{\partial x}\right|_{x=-5} = \left.\frac{\partial u}{\partial x}\right|_{x=5}$$

Because of the complexity of these boundary conditions, the policy network is set as $\hat{u}(x,t) = f(x,t;\theta)$, not as that used for Burgers' equation. A DNN consisting of seven hidden layers is used for inferring the policy. The architecture of the hidden layers is the same as that used for solving Burgers' equation.

The critic for each time step is

$$r = r_B + r_I + r_{Eq},$$

$$r_B = -\frac{1}{N_B} \sum \left(\hat{u}(x_B,t) - \hat{u}(x_{-B},t)\right)^2 \bigg|_{x_B = 5, x_{-B} = -5},$$

$$r_I = -\frac{1}{N_I} \sum \left(\hat{u}(x_I,0) - u(x_I,0)\right)^2, \tag{13}$$

$$r_{Eq} = -\frac{1}{N_{Eq}} \sum \left(\frac{\partial \hat{u}}{\partial t} - 0.5i \frac{\partial \hat{u}}{\partial x^2} - i|\hat{u}|^2 \hat{u}\right)^2,$$

where $N_B, N_I, N_{Eq}$ are the numbers of sampled points for the boundary condition, initial condition, and equation, respectively. The loss function can then be calculated using Eq. (2).

The temporal interval $\Delta t$ is set as 0.01, and the input of the policy network of each discretized step $i$ is a minibatch of $s_i = (t_i, x_i)$. The output of the network includes the real and imaginary parts of the solution $\mu_\theta(s_i) = [real(\mu_u) \; image(\mu_u)]$ and the standard deviation $[real(\sigma_u) \; image(\sigma_u)]$. The learning rate is set to exponential decay, with an initial value of 5 × 10$^{-4}$ and a decay rate of 0.95 (decaying every 100 steps, but no less than 3 × 10$^{-6}$). The numbers of spatial–temporal points sampled in this example are $N_B = 1{,}000$, $N_I = 100$ and $N_{Eq} = 999$, respectively, and $999i$ points (less than 20,000) are sampled for the previous step to form the $s_i = (t_i, x_i)$. After iteration, the mean square of these three critics converges to less than $5 \times 10^{-5}$ for each time step. Comparisons between the DRL solutions and numerical solutions obtained by a high-order numerical method, namely, the fourth-order Runge–Kutta exponential time differencing method (ETD4RK) [35, 36], are shown in Fig. 4. Both the real and imaginary parts of the DRL solution, i.e., the spatial–temporal cloud map of the solution and the solutions selected at some spatial points, agree well with the ETD4RK solution.

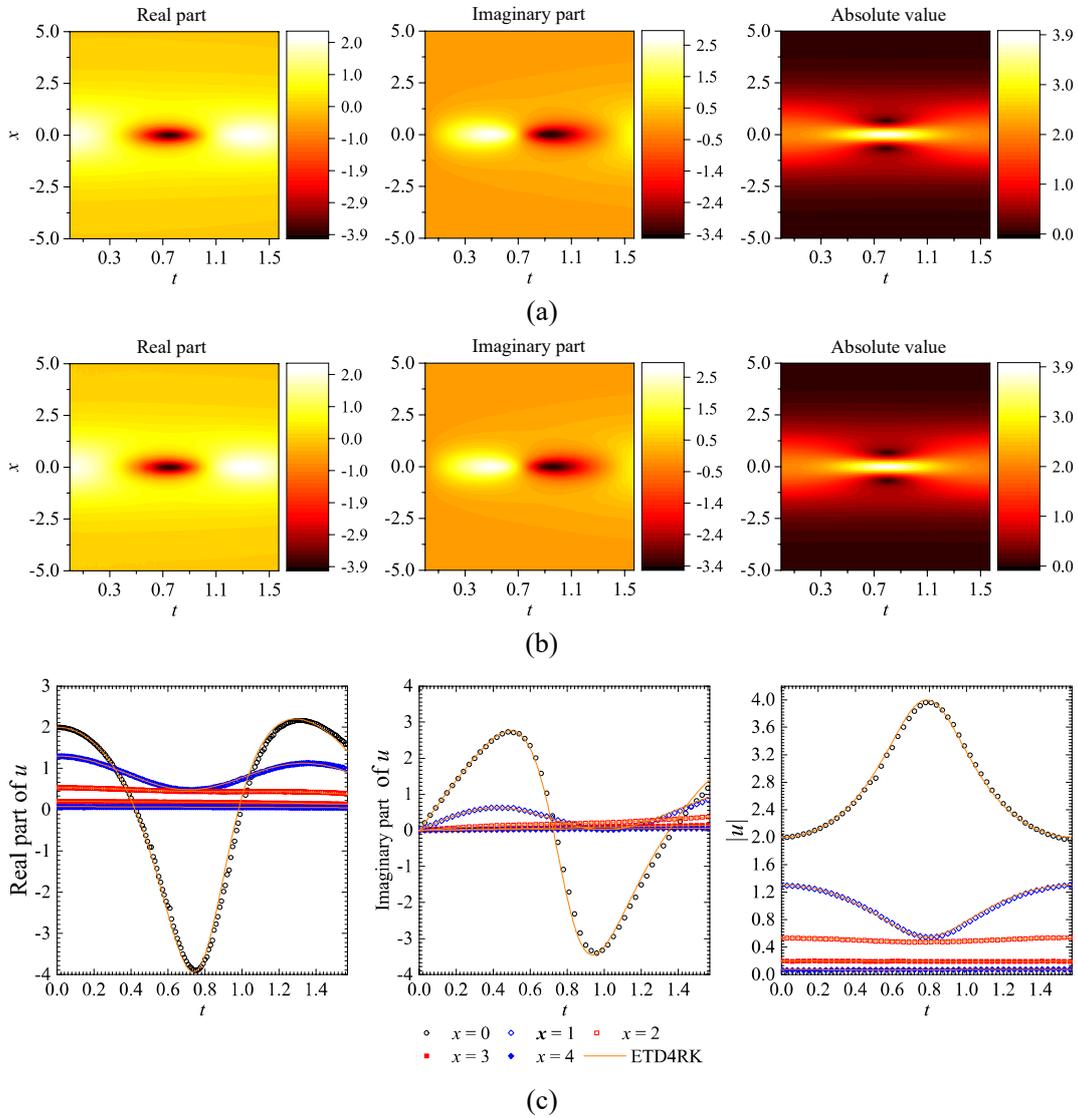

**Fig. 4** Solutions of the one-dimensional Schrödinger equation: (a) spatial–temporal ETD4RK solution; (b) spatial–temporal DRL solution; (c) comparison between the DRL and ETD4RK solutions at five selected spatial points.

## 3.6 Navier-Stokes equations

The Navier–Stokes equations of fluid dynamics are

$$\frac{\partial \mathbf{u}}{\partial t} + (\mathbf{u} \cdot \nabla)\mathbf{u} = -\frac{1}{\rho}\nabla p + \frac{\mu}{\rho}\nabla^2 \mathbf{u} \tag{14}$$
$$\nabla \cdot \mathbf{u} = 0$$

where $\mathbf{u}(\mathbf{x},t)$ is the velocity field, $p(\mathbf{x},t)$ is the pressure, $\rho$ is the density of the fluid and $\mu$ is the viscosity coefficient. To test the possible application of the DRL approach to fluid dynamics, the steady Couette flow in Fig. 5a is taken as an example. The governing equations are formulated as,

$$(\mathbf{u} \cdot \nabla)\mathbf{u} = -\frac{1}{\rho}\nabla p + \frac{\mu}{\rho}\nabla^2 \mathbf{u}$$
$$\nabla \cdot \mathbf{u} = 0$$

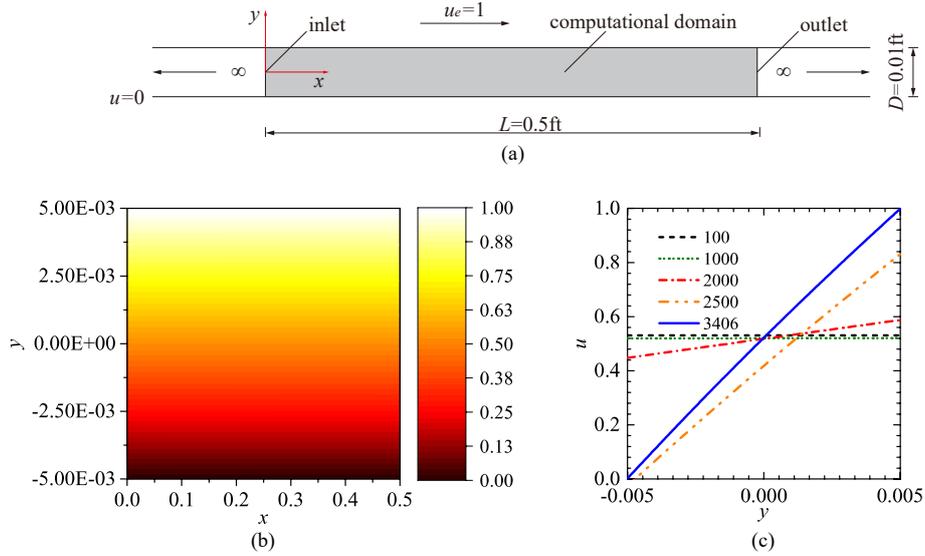

**Fig. 5** Couette flow: (a) Computational domain of the flow; (b) spatial–temporal DRL solution; (c) streamwise velocity at $x = 0.25$ for various iteration steps.

The critic for a single time step is

$$r = \lambda r_B + r_{Eq},$$

$$r_B = -\frac{1}{N_B}\sum \left(\hat{\mathbf{u}}(\mathbf{x}_B) - \mathbf{u}(\mathbf{x}_B)\right)^2, \tag{15}$$

$$r_{Eq} = -\frac{1}{N_{Eq}}\sum\left((\hat{\mathbf{u}}\cdot\nabla)\hat{\mathbf{u}} + \frac{1}{\rho}\nabla p - \frac{\mu}{\rho}\nabla^2\hat{\mathbf{u}}\right)^2 - \frac{1}{N_{Eq}}\sum(\nabla\cdot\hat{\mathbf{u}})^2,$$

where $N_B, N_{Eq}$ are the numbers of sampled points for the boundary and initial conditions and the equation, and $\lambda$ is a hyperparameter preventing the effects of imbalanced sampling. The loss function can then be calculated using Eq. (2).

The computational domain and the initial and boundary conditions for the Navier–Stokes equations are as follows:

$$x \in [0, 0.5], y \in [-0.005, 0.005]$$
$$u=1|_{y=0.005}, \ v=0|_{y=0.005}, \ u=v=0|_{y=-0.005}$$
$$p=0|_{x=0}, \ v=0|_{x=0}, \ p=0|_{x=0.5}$$

Three policy networks are used in the DRL framework:

$$\begin{aligned}\hat{u}(x,y) &= f_u(x,y;\theta_u) \\ \hat{v}(x,y) &= f_v(x,y;\theta_v) \\ \hat{p}(x,y) &= f_p(x,y;\theta_p)\end{aligned} \quad (16)$$

where $\theta_\mu$, $\theta_v$ and $\theta_p$ are the trainable parameters of the three networks. A DNN consisting of seven hidden layers is used for inferring the policy. The numbers of nodes in each hidden layer are [20 30 40 40 40 30 20], and the activation function is tanh. The numbers of spatial–temporal points sampled in this example are $N_b = 4,000$, $N_i = 2,000$, and $N_e = 20,000$. The learning rate is set to exponential decay, with an initial value of $10^{-3}$ and a decay rate of 0.995 (decaying every 15 steps, but no less than 2 × $10^{-6}$). It should be noted that the solutions of this PDE is governed by its boundary conditions. Hence, $\lambda$ is set to large values at the beginning to accelerate the training process, and exponential decay is applied to this hyperparameter. The exponential decay is applied to $\lambda$ corresponding to the upper and lower boundary conditions with an initial value of 50 and a decay rate of 0.995 (decaying every 15 steps, but no less than 1), and $\lambda$ corresponding to the inlet and outlet boundary conditions is set to 1 during the training process. After iterations, the mean square error converges to less than 5 × $10^{-5}$. The DRL solution is shown in Fig. 5b. The DRL approach gives the correct velocity distribution of the Couette flow. Figure 5c shows the convergence process of this method.

## 4. Discussions

### 4.1 Transfer learning characteristics

During the training of the DRL network, solutions in discretized time are treated as multiple tasks that share the same governing equation, and because the solutions are temporally continuous, the parameters of the policy network at the current time step provide an ideal initialization for the next time step, behaving as transfer learning characteristics. Consequently, training may become faster and faster, so that for some initial conditions, the solution converges after only one or few steps. For instance, it is very interesting that the iteration step number decreases as solving process proceeds, as shown in Fig. 6, which may imply that the training network has grasped the intrinsic properties of the equations in these cases. It can be seen that the time cost decays with time step in general, which indicates that the DRL solver promises to "learn" the solution during the solving procedure.

Figure 7 gives the mean square loss versus iteration step for the Van der Pol and Burger's equations for some selected time steps. It can be seen that while the initial mean square loss for time step 1 is high, the initial mean square losses are much lower for later time steps by several orders of magnitude. For $v = 0.05$ and $v = 0.01$, the mean square loss for step 100 may be large at the first steps, but it decreases rapidly to small values. In addition, the time cost decays with time step in general, also indicating that the DRL solver promises to "learn" the solution during the solving procedure without the need of much knowledge in mathematics about ODEs and PDEs. Furthermore, cumbersome grids of high quality are not needed in the DRL approach, which thus differs from the traditional finite difference and finite volume methods. Moreover, the most important contribution of this study lies in the provision of a general rule-based solution framework to solve differential equations by DRL, such that people (or AI agents) can learn to solve any type of differential equation with little or no background knowledge of the analytical characteristics of the equation. This general and simple structure solver will furthermore lead to a more intelligent agent that is able to better understand the physical world.

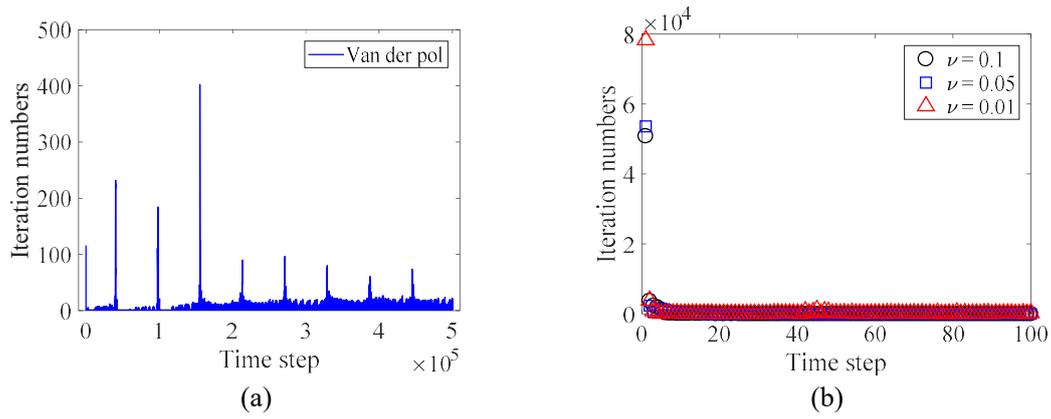

**Fig. 6** Iteration step versus time step for (a) the Van der Pol equation and (b) Burgers' equation.

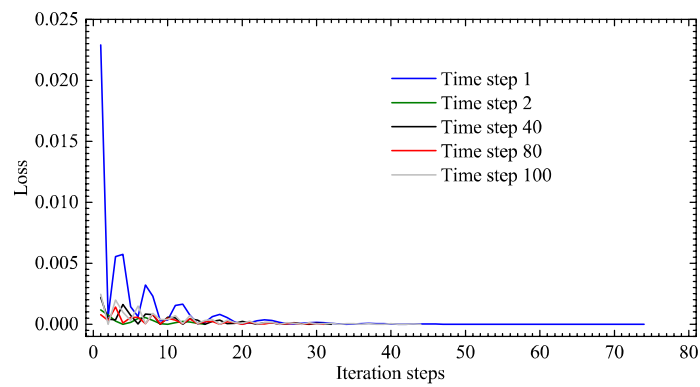

(a)

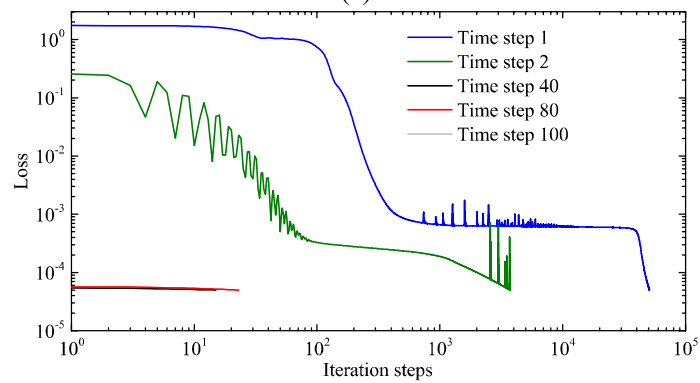

(b)

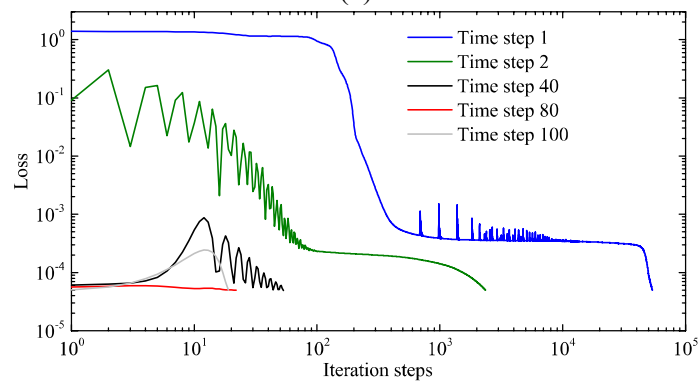

(c)

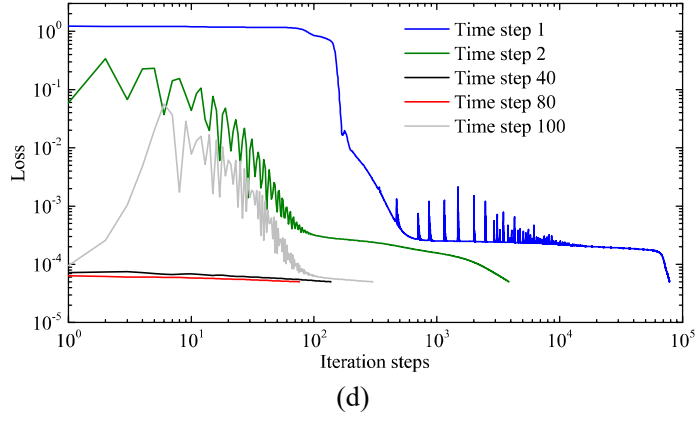

(d)

**Fig. 7** Mean square loss versus iteration step for (a) the Van der pol equation and (b–d) Burger's equation with (b) $v = 0.1$, (c) $v = 0.05$, and (d) $v = 0.01$.

## 4.2 Limitations

The DRL solver fails to satisfy the symmetric property of the Lorenz equation when the system becomes chaotic, because the chaotic system is very sensitive to initial states, i.e., the solving processes for different initial states experience different training procedures, and this causes tiny rounding errors in the solution. However, even a tiny rounding error will induce quite large differences in the solution after a long time. Taking the Lorenz equations for example, four Rayleigh numbers, ranging from 15 to 28, are considered (as shown in Fig. 8). Nonzero values of $x_1(t) + x_2(t)$, $y_1(t) + y_2(t)$ or $z_1(t) - z_2(t)$ indicate that the DRL solutions deviate from the theoretical values when the Rayleigh number is larger than a threshold value. It is obvious that the DRL solver successfully captures the symmetric property when the system is stable ($\rho = 15$ and 24); however, it fails for the chaotic systems ($\rho = 26$ and 28). The root mean squares of $x_1(t) + x_2(t)$, $y_1(t) + y_2(t)$, and $z_1(t) - z_2(t)$ are illustrated in Fig. 8 to show the computational error of the DRL solver and identical conclusions can be draw.

Moreover, inspired by the trial-and-error mechanism, the DRL agent learns from the trial-and-error experience to obtain the solution of the equation, thus requiring more trial-and-error experience and a larger and deeper network (thus longer training time) for large-domain problems. Therefore, the DRL approach involves the training of larger and deeper networks, which is the most expensive part of the process, and can incur a greater time cost compared with regular numerical methods. Although it is with no superior performance than the existing numerical methods with regard to computational cost for the current stage, the DRL approach does enable the computer agent to directly self-learn to solve any type of equation based on the given rules without the need of much knowledge in mathematics about ODEs and PDEs. Therefore, this approach may shed light on self-learning agents (or machines) in the scientific computing community.

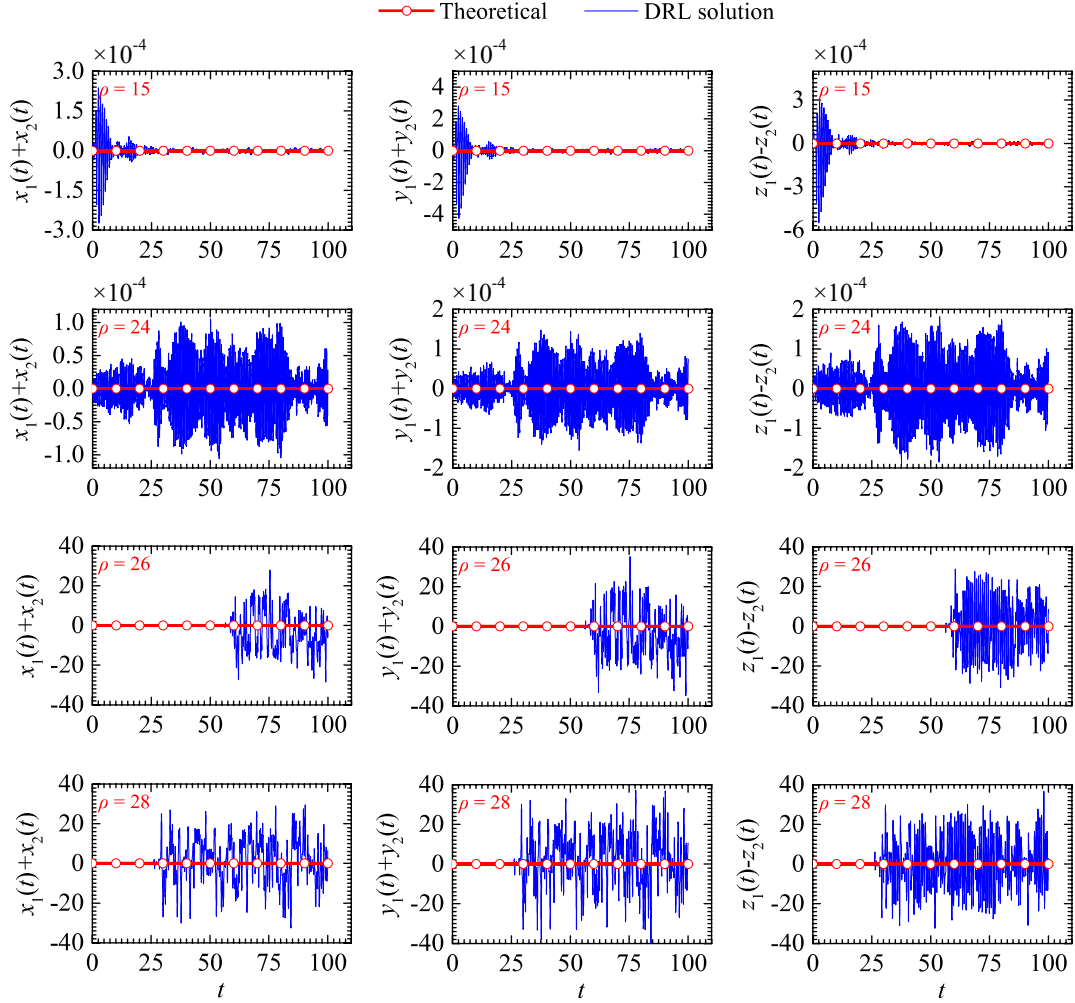

**Fig. 8** Capture of the symmetric property of the Lorenz equations for various Rayleigh numbers: the column on the left represents the time series of $x_1(t)+x_2(t)$, the column on the middle the time series of $y_1(t)+y_2(t)$, and the column on the right the time series of $z_1(t)-z_2(t)$.

# 5. Conclusions

In summary, we have presented a novel rule-based DRL approach for solving differential equations without any data or any built-in knowledge about the equations, and the equation is embedded in the network as the critic, which makes sense from a physical point of view.

The DRL solver consists of a DNN structured actor that outputs policy to approximate the solutions, together with a rule-based critic that evaluates the output solution of the actor. The policy network of the actor provides a nonlinear mapping from the computational domain to the solution domain and is trained in the batch method, by inputting the sampling points from the computational domain and outputting the mean and variance to construct the probabilistic policy in Gaussian form. The candidate solution is sampled from the output policy, in order to preserve the gradient propagation for accelerating the training, and the candidate solution is sampled as the output mean value. The physical governing equation (general form, including boundary and initial conditions) acts as the critic to indicates the balance of the equation when the candidate solution is substituted into it. The gradients of the policy network are calculated based on the critic. All the DRL solutions of these differential equations agree well with those obtained by high-order numerical methods and with analytical solutions. In particular, this initial study shows the potential of the DRL approach to solve the Navier–Stokes equations, although only a steady Couette flow case has been tested.

An exact solution of any differential equation can always be obtained if the loss function converges to a small value; i.e., this method provides a general way to solve various differential equations. After careful verification of the proposed approach, problems without analytical or numerical solutions will be studied in the future, which will make the method even more impactful.

However, because the last-step solution is used as the initial condition for the next step, the proposed method could fail for equations that are sensitive to initial values (the effects of accumulated errors may be larger than the effects of the initial values), such as in the case of chaotic systems exhibiting bifurcations. It should also be mentioned that at its current stage of development, the DRL solver fails for complex problems in which the objective function has a number of local optima, such as turbulence problems described by the Navier–Stokes equations. We intend in the future to develop the approach to exploit the low-dimensional features of a system, as has already been investigated in the literature [37], since this may simplify the training process for such complex problems.

## Acknowledgement

This research was funded by the National Natural Sciences Foundation of China (NSFC) (Grant Nos. U1711265 and 51638007).